\documentclass[10pt,twocolumn,letterpaper]{article}

\usepackage[pagenumbers]{cvpr} 

\usepackage{amsmath,amsfonts}
\usepackage{array}
\usepackage{textcomp}

\usepackage{verbatim}
\usepackage{dsfont}
\usepackage{booktabs, multirow}
\usepackage{amssymb}
\usepackage{pifont}
\usepackage{tabularx}
\usepackage{float}
\usepackage{stfloats}
\usepackage{bm}

\usepackage[accsupp]{axessibility} 

\definecolor{cvprblue}{rgb}{0.21,0.49,0.74}
\definecolor{lightlightgray}{rgb}{0.902, 0.902,0.902}
\usepackage[pagebackref,breaklinks,colorlinks,allcolors=cvprblue]{hyperref}

\def\paperID{16361} 
\def\confName{CVPR}
\def\confYear{2025}

\title{Pursuing Temporal-Consistent Video Virtual Try-On \\via Dynamic Pose Interaction}

\author{Dong Li$^{1}$, Wenqi Zhong$^{2 \ast}$, Wei Yu$^{1}$, Yingwei Pan$^{1}$, Dingwen Zhang$^{2}$, \\Ting Yao$^{1}$, Junwei Han$^{2}$, and Tao Mei$^{1}$\\
\parbox{40em}{\small\centering $^{1}$ HiDream.ai Inc. ~~~~~~~~~~~~$^{2}$ Northwest Polytechnical University, Xi'an, China}\\
{\tt\small lidong@hidream.ai, wenqizhong@mail.nwpu.edu.cn, \{yuwei, pandy\}@hidream.ai} \\
{\tt\small zhangdingwen2006yyy@gmail.com, tiyao@hidream.ai, junweihan2010@gmail.com} \\
{\tt\small tmei@hidream.ai}
}

\begin{document}
\maketitle
\let\thefootnote\relax\footnotemark\footnotetext{$^{\ast}$ Equal contribution. This work was performed at HiDream.ai.}
\begin{abstract}

Video virtual try-on aims to seamlessly dress a subject in a video with a specific garment. The primary challenge involves preserving the visual authenticity of the garment while dynamically adapting to the pose and physique of the subject. While existing methods have predominantly focused on image-based virtual try-on, extending these techniques directly to videos often results in temporal inconsistencies. Most current video virtual try-on approaches alleviate this challenge by incorporating temporal modules, yet still overlook the critical spatiotemporal pose interactions between human and garment. Effective pose interactions in videos should not only consider spatial alignment between human and garment poses in each frame but also account for the temporal dynamics of human poses throughout the entire video. With such motivation, we propose a new framework, namely \textbf{D}ynamic \textbf{P}ose \textbf{I}nteraction \textbf{D}iffusion \textbf{M}odels (DPIDM), to leverage diffusion models to delve into dynamic pose interactions for video virtual try-on. Technically, DPIDM introduces a skeleton-based pose adapter to integrate synchronized human and garment poses into the denoising network. A hierarchical attention module is then exquisitely designed to model intra-frame human-garment pose interactions and long-term human pose dynamics across frames through pose-aware spatial and temporal attention mechanisms. Moreover, DPIDM capitalizes on a temporal regularized attention loss between consecutive frames to enhance temporal consistency. Extensive experiments conducted on VITON-HD, VVT and ViViD datasets demonstrate the superiority of our DPIDM against the baseline methods. Notably, DPIDM achieves VFID score of 0.506 on VVT dataset, leading to 60.5\% improvement over the state-of-the-art GPD-VVTO approach.

\end{abstract}    
\begin{figure}[t]
\centering
	\setlength{\abovecaptionskip}{0pt}
    \includegraphics[width=3.0in]{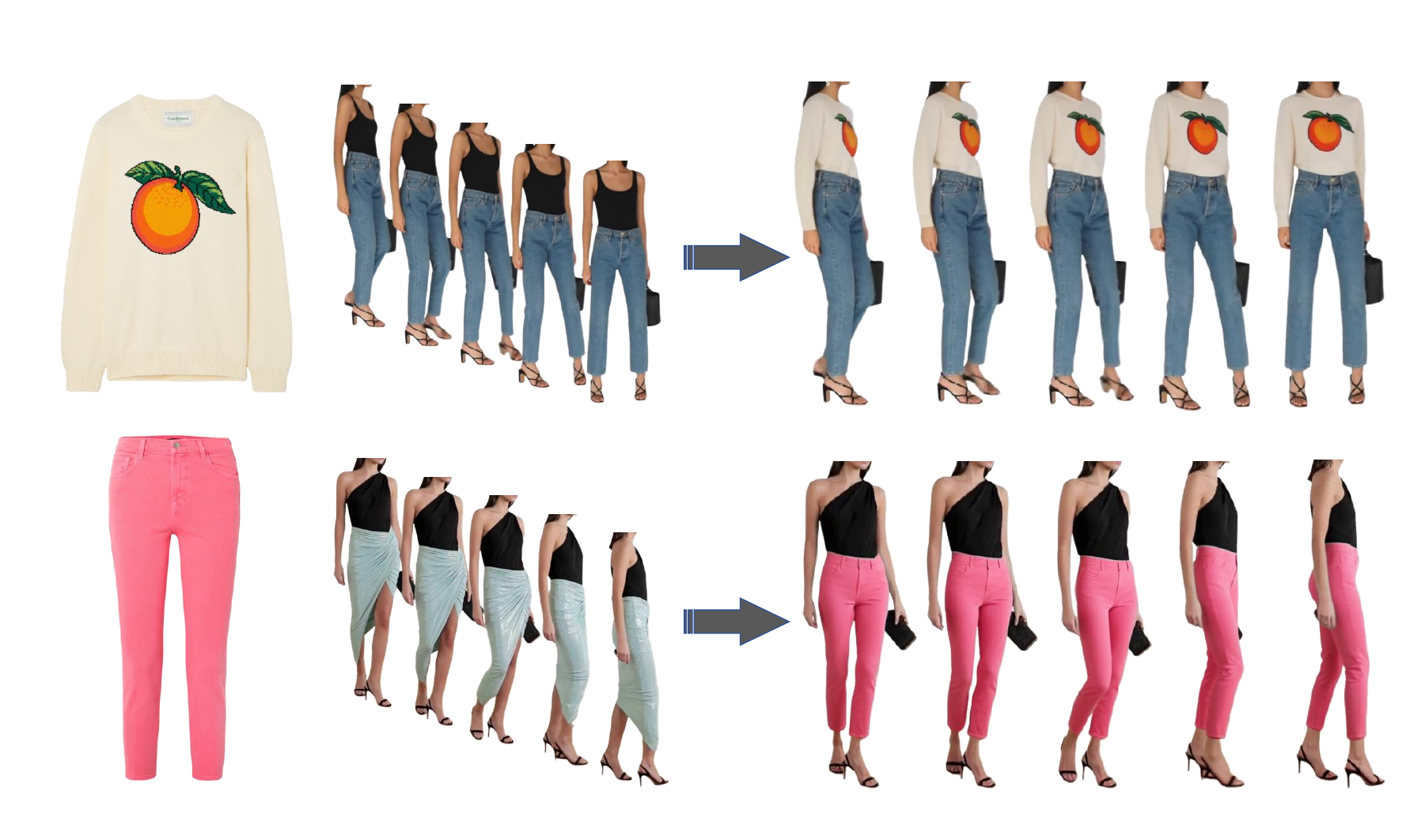}
    \caption{Given a garment and a person video, DPIDM generates a lifelike video that preserves the visual authenticity of the garment while adapting to the individual's pose and physique.}
    \label{fig:introduction}
\vspace{-0.2in}
\end{figure}

\section{Introduction}
\label{sec:intro}
Generative models have made tremendous progress in recent years, with wide-ranging applications across various fields, including computer vision \cite{pan2017create, croitoru2023diffusion, zhang2024trip}, computer graphics \cite{chen2024text,gao2024guess}, aesthetic design \cite{oh2019deep}, and medical research \cite{hamamci2025generatect}. Virtual try-on is a quintessential generative task that aims to synthesize a lifelike image/video of the specific person wearing the provided garment. By harnessing the powerful generative capacities of diffusion models, this field has emerged as a promising domain with versatile applications in immersive e-commerce and short-form video platforms.

Prior research mainly focused on image-based virtual try-on. The earlier approaches typically build on Generative Adversarial Networks (GANs) \cite{han2018viton, wang2018toward, choi2021viton, rombach2022high}, incorporating a warping module alongside a try-on generator. The warping module deforms clothing to align with the human body, and then the warped garment is fused with the person image through the try-on generator. However, in light of the recent emergence of UNet-based Latent Diffusion Models (LDMs) \cite{xu2024ootdiffusion, zhu2023tryondiffusion, choi2024improving, kim2023stableviton, chen2024wear}, researchers have progressively directed their focus towards these innovative models. A diffusion-based try-on network combines the warping and blending processes into a unified operation of cross-attention without explicit segregation. By employing pre-trained text-to-image weights, these diffusion approaches demonstrate superior fidelity compared to their GAN-based counterparts.

In recent years, researchers have endeavored to replicate the success of image try-on within the realm of video. One direct approach involves applying image try-on techniques to process videos frame by frame. However, this method often results in notable inter-frame inconsistencies. To address this challenge, various specialized designs have been explored for video virtual try-on \cite{dong2019fw, jiang2022clothformer, zhong2021mv, kuppa2021shineon}. These approaches commonly integrate optical flow prediction modules to warp frames produced by the try-on generator. Despite reasonable performance, GAN-based methods encounter difficulties in handling garment-person misalignment, particularly when faced with inaccurate warping flow estimations. To harness the capabilities of pre-trained image-based diffusion models, TunnelTry-on \cite{xu2024tunnel} and ViViD \cite{fang2024vivid} utilize a pre-trained inpainting U-Net as the primary branch and integrate a reference U-Net to capture detailed clothing features. They enhance temporal consistency by incorporating standard temporal attention after each stage of the main U-Net. GPD-VVTO \cite{wang2024gpd} takes a step further by integrating garment features into temporal attention, enabling the model to better preserve fidelity to the garment during temporal modeling. Despite showing promise, these methods struggle to maintain both visual integrity and motion consistency simultaneously, particularly when there is a significant stylistic gap between the garment being tried on and the original one. To alleviate these issues, we argue that two key ingredients should be taken into account. One is spatial alignment between human and garment poses within each frame. Such alignment is essential for virtual try-on as it ensures that clothing adapts appropriately to factors like coverage and wrinkles based on the individual's posture. The other is the temporal dynamics of human poses across the entire video. A comprehensive understanding of this long-term interaction is paramount for upholding temporal coherence, given that human movements naturally progress as a continuum of interconnected poses.

By consolidating the idea of modeling both spatial and temporal pose interactions in a video, we novelly present \textbf{D}ynamic \textbf{P}ose \textbf{I}nteraction \textbf{D}iffusion \textbf{M}odels (DPIDM) for boosting video virtual try-on. Specifically, DPIDM employs a dual-branch architecture, where the Main U-Net conducts denoising processes on the video, while the Garment U-Net extracts intricate details from the garment image and integrates them into the Main U-Net. This architecture facilitates a holistic understanding of the relationship between humans and garments within a unified feature space. To effectively leverage the spatiotemporal pose interactions in videos, we introduce a skeleton-based pose adapter that seamlessly integrates synchronized human and garment poses into the Main U-Net. With the adapter, we present a hierarchical attention module comprising four key components: a pose-aware spatial attention block, a temporal-shift attention block, a cross-attention block, and a pose-aware temporal attention block. The pose-aware spatial attention block plays a crucial role in capturing intricate intra-frame human-garment pose interactions, while the pose-aware temporal attention block models the long-term dynamics of human poses across frames. Moreover, a temporal regularized attention loss is defined on consecutive frames to ensure temporal coherence. As a result, our DPIDM excels in generating videos with meticulously aligned garment details and enhanced temporal stability.

The main contribution of this work is the proposal of Dynamic Pose Interaction Diffusion Models that facilitate the video virtual try-on (VVTON) task. This also leads to the elegant view of how a diffusion model should be designed for excavating the pose-focused prior knowledge (e.g., spatial pose alignment and temporal pose dynamics) tailored to VVTON, and how to improve diffusion process with these amplified pose-focused guidance. Through an extensive set of experiments on VITON-HD, VVT, and ViViD datasets, DPIPM consistently achieves superior results over state-of-the-art methods in both image and video virtual try-on tasks.

\section{Related Works}
\subsection{Image-based Virtual Try-on}
The development of image-based virtual try-on is largely inspired by the success of GANs \cite{goodfellow2014generative}, leading to pioneering works such as \cite{han2018viton, wang2018toward, dong2020fashion, morelli2022dress, choi2021viton, li2023virtual}. A standard GAN-based virtual try-on framework typically adheres to a two-stage paradigm. The initial stage involves an image warping process, such as Thin Plate Spline (TPS) warping \cite{han2018viton}, which aligns the garment with the target individual's body. In the second stage, a GAN-based try-on generator merges the warped attire with the target individual's clothing-agnostic representation to produce authentic try-on outcomes. However, these methods heavily rely on the warping module to preserve garment details and often suffer from performance degradation due to inaccurate warping.

Recently, with the powerful generative capabilities of Latent Diffusion Models (LDMs), many diffusion-based models have been introduced to generate more realistic virtual try-on results \cite{kim2023stableviton, zeng2024cat, yang2024texture, zhu2023tryondiffusion, chen2024wear, wan2024improving}. TryOnDiffusion \cite{zhu2023tryondiffusion} was the first to propose a dual U-Net architecture that simultaneously preserves garment details and implicitly warps the garment to match the target pose. StableVITON \cite{kim2023stableviton} introduced a zero cross-attention block to learn the semantic correspondence between the garment and the target person. Wear-any-way \cite{chen2024wear} utilized a sparse point control method to align person and garment features in feature space. MMTryon \cite{zhang2024mmtryon} presented a multi-modal, multi-garment virtual try-on approach that combines text and multiple reference images to create a multimodal embedding as a condition to control the diffusion model. IDM \cite{choi2024improving} proposed a RefNet-style architecture to align the feature space between garment and person. Although these diffusion-based methods have achieved high-fidelity results in single-image inference, their application to video virtual try-on reveals a crucial limitation: the oversight of inter-frame relationships. This deficiency results in notable inter-frame inconsistencies, ultimately yielding unsatisfactory outcomes.


\begin{figure*}[t]
\centering
	\setlength{\abovecaptionskip}{0pt}
    \includegraphics[width=0.96\linewidth]{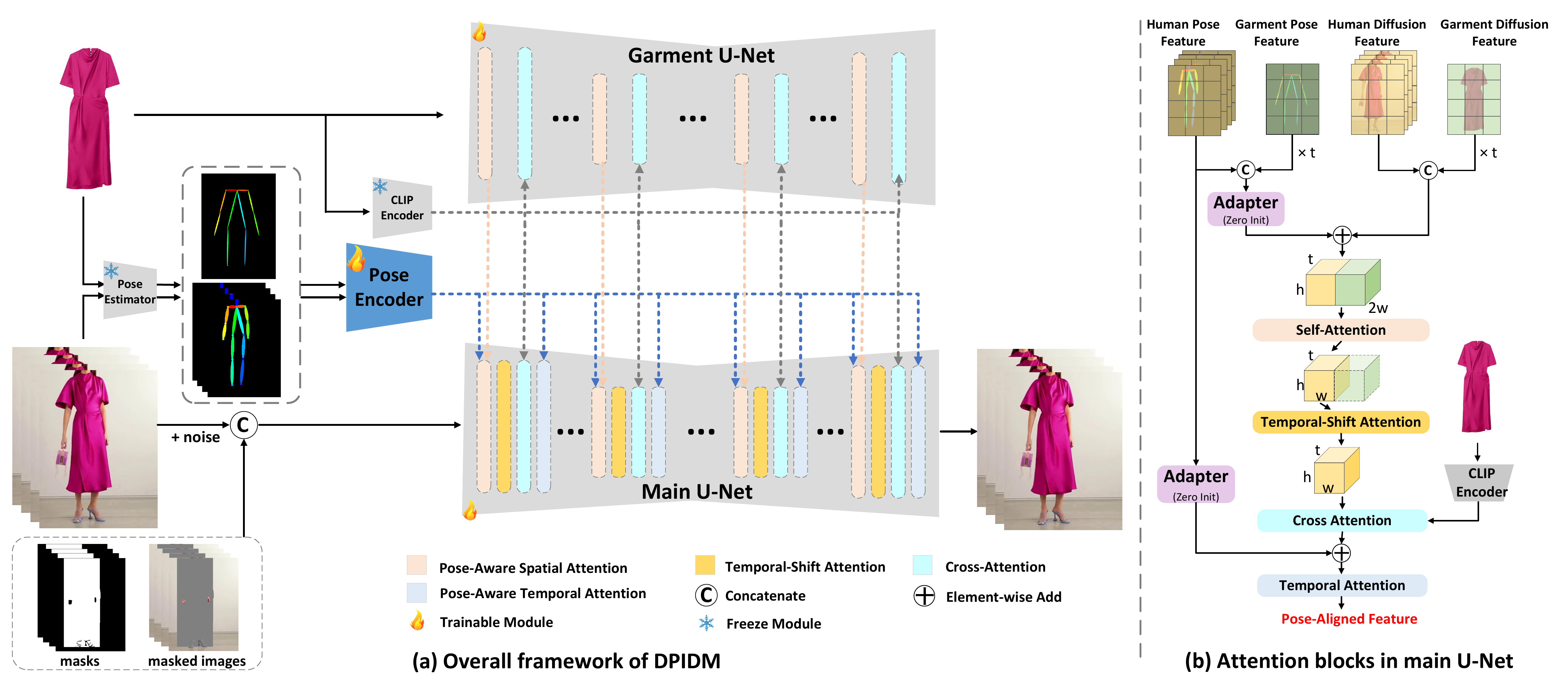}
    \caption{(a) Overall architecture of DPIDM. DPIDM emplys a dual-branch architecture. The main U-Net processes a concatenated input comprising the noisy latent of the video, the latent of the cloth-agnostic video, and the cloth-agnostic mask sequence. The garment U-Net extracts fine-grained garment features, which are subsequently integrated into the main U-Net. The pose estimator is utilized to extract aligned human and garment poses, which are then fed into the attention modules of the main U-Net to guide the diffusion process. The VAE is not shown for clarity. (b) Detailed illustration of the proposed pose-aware attention module within the main U-Net. The module comprises pose-aware spatial attention, temporal-shift attention, cross-attention, and pose-aware temporal attention. The pose embeddings are seamlessly integrated into the attention module through a specialized pose adapter. (better viewed in color)}
    \label{fig:framework}
\vspace{-0.18in}
\end{figure*}

\subsection{Video-based Virtual Try-on}
In recent studies, researchers have aimed to expand the capabilities of image-based try-on approaches into video applications. A key issue in video generation is inter-frame inconsistency. ClothFormer \cite{jiang2022clothformer} leverages optical flow in the warping and blending processes to achieve temporal smoothing. Recently, diffusion-based video virtual try-on methods have been introduced to generate more realistic results \cite{xu2024tunnel, fang2024vivid, wang2024gpd, he2024wildvidfit, zheng2024viton-dit}. For instance, TunnelTry-on \cite{xu2024tunnel} proposed a focus-tunnel method that crops the human image according to the pose map, aligning the try-on image to accommodate scenarios where the person is off-center in the frame. VTON-DiT \cite{zheng2024viton-dit} introduced a DiT architecture compatible with OpenSora, where the clothing-agnostic human image is provided via ControlNet. WildVidFit \cite{he2024wildvidfit} proposed an image-based diffusion model and used video MAE to update the intermediate representations of the diffusion process to improve frame consistency. ViViD \cite{fang2024vivid} introduced a large-scale video try-on dataset and proposed an additional reference-based U-Net. GPD-VVTON \cite{wang2024gpd} went a step further by integrating garment features into temporal attention to generate spatio-temporally consistent results. While these explorations of video try-on make steady advancements, they currently struggle to accurately capture the dynamic interaction between clothing and the human body in real-world scenarios. Our method employs a hierarchical attention module to capture the critical spatiotemporal pose interactions between the human and the garment.

\section{Methodology}
In this paper, we present \textbf{D}ynamic \textbf{P}ose \textbf{I}nteraction \textbf{D}iffusion \textbf{M}odels (DPIDM), a pioneering framework based on latent diffusion models designed to enhance the VVTON task. The complete architecture is depicted in Figure \ref{fig:framework} (a). This section begins with a brief introduction to Stable Diffusion \cite{rombach2022high} in Section \ref{sec:preliminary} to lay the groundwork for subsequent discussions. Subsequently, we offer a comprehensive overview of DPIDM in Section \ref{sec:overview}, and delve into the specific design details in Section \ref{sec:pose}-\ref{sec:loss}. Finally, we encapsulate our training and inference strategies in Section \ref{sec:train}.

\subsection{Preliminary} \label{sec:preliminary}
Stable Diffusion (SD) stands out as one of the most prevalent latent diffusion models within the community. It leverages a variational autoencoder (VAE) architecture, comprising an encoder $\mathcal{E}$ and a decoder $\mathcal{D}$, to facilitate image representations in the latent space. Additionally, a U-Net~\cite{ronneberger2015u} $\epsilon_{\theta}$ is trained to denoise a Gaussian noise $\epsilon$ with a conditioning input encoded by a CLIP text encoder~\cite{radford2021learning} $\tau_{\theta}$. Given an image $\mathbf{x}$ and a text prompt $\mathbf{y}$, the denoising U-Net $\epsilon_{\theta}$ undergoes training by minimizing the loss function:
\vspace{-0.05in}
\begin{equation} \label{eq:ldm}
    \mathcal{L}_{LDM} = \mathbb{E}_{\mathcal{E}(\mathbf{x}),\mathbf{y},\epsilon\sim\mathcal{N}(0, 1),t}\left[\lVert\epsilon - \epsilon_{\theta}(\mathbf{z}_t, t, \tau_{\theta}(\mathbf{y}))\rVert_2^2\right],
\end{equation}
where $t$ denotes the time step of the forward diffusion process, and $\mathbf{z}_t$ is the noisy latent constructed by adding Gaussian noise $\epsilon\sim\mathcal{N}(0, 1)$ to the encoded image $\mathcal{E}(\mathbf{x})$. During the inference stage, $\mathbf{z}_t$ is randomly sampled from a Gaussian distribution and iteratively denoised to derive $\mathbf{z}_0$ following a predefined sampling schedule \cite{ho2020denoising, song2020denoising}. Finally, the latent decoder $\mathcal{D}$ decodes $\mathbf{z}_0$ back into the image space.

\subsection{Overview} \label{sec:overview}
The overall architecture of DPIDM is depicted in Fig. \ref{fig:framework}. DPIDM takes a reference garment image $\boldsymbol{G} \in \mathbb{R}^{3 \times H \times W}$ and a source human video $\boldsymbol{I}_{S} \in \mathbb{R}^{T \times 3 \times H \times W}$ as input, where \emph{H} and \emph{W} represent the height and width of each frame, and \emph{T} signifies the video length. Its objective is to generate a realistic video $\hat{\boldsymbol{I}} \in \mathbb{R}^{T \times 3 \times H \times W}$ showing the individual in $\boldsymbol{I}_{S}$ adorned in the garment $\boldsymbol{G}$. 

We approach the video virtual try-on as a video inpainting challenge, aimed at integrating the garment onto clothing-agnostic areas. To achieve this, DPIDM employs a dual-branch architecture. The main U-Net is the inpainting model initialized with the pre-trained weights from SD. It takes in a 9-channel tensor comprising 4 channels for the noisy latent of the video, 4 channels for the clothing-agnostic video latent, and 1 channel for the binary agnostic mask sequence. In contrast to the original SD model, which utilizes text embeddings to guide the diffusion process, we have substituted these embeddings with image embeddings derived from a CLIP image encoder representing the garment image. While the CLIP image embedding effectively captures the overall colors and textures of the garment, it may not retain finer details. To address this, we also use a Garment U-Net to extract detailed garment features. The Garment U-Net follows a standard text-to-image diffusion model with a 4-channel input. This dual-branch architecture has proven to be effective by previous works \cite{fang2024vivid, xu2024tunnel, wang2024gpd}.

To further enhance the generation, we incorporate human and garment poses as supplementary guidance to refine the diffusion process. We develop a light-weight pose encoder to extract features of the pose maps and a pose adapter to inject these features into the attention modules of the main U-Net. The human/garment pose maps are derived from the input video/image with pose estimator.

\subsection{Pose Estimator} \label{sec:pose}
\begin{figure}[t]
\centering
    \setlength{\abovecaptionskip}{0pt}
    \includegraphics[width=3.2in]{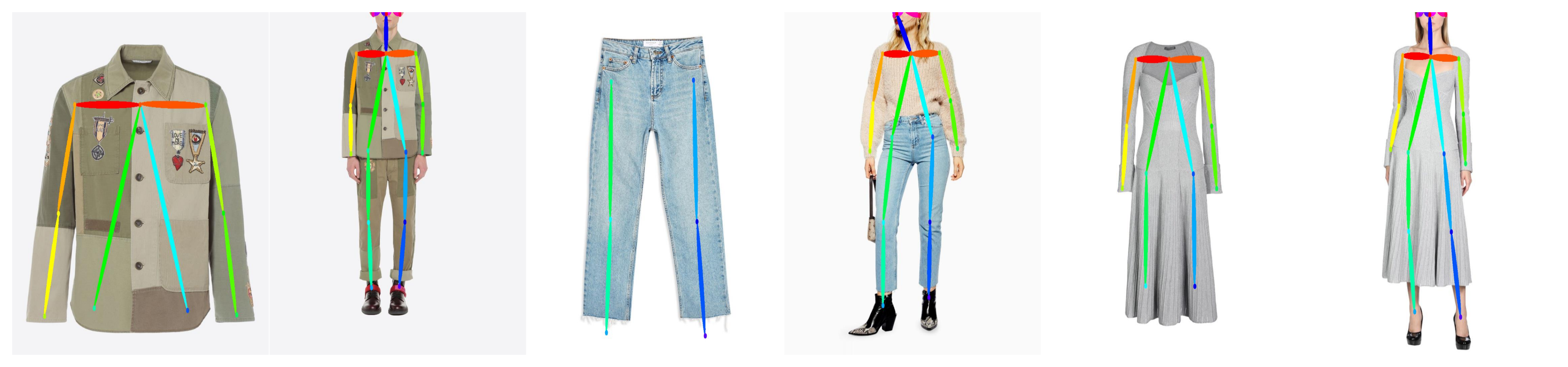}
    \caption{Visualization of predicted human and garment poses.}
    \label{Visual_image}
\label{fig:pose}
\vspace{-0.18in}
\end{figure}

We directly employ DW-Pose \cite{yang2023effective} to estimate human poses $\bm{P}_h$ from video frames. However, there is a lack of an open-source garment pose estimator currently. To establish alignment between human and garment poses, we train a garment pose estimator to predict a set of landmarks $\bm{P}_g$ that correspond to various parts of the human pose $\bm{P}_h$. The ground truth of $\bm{P}_g$ is labeled manually, with the number of landmarks varying based on the garment types. For instance, in upper clothing, $|\bm{P}_g|$=9, encompassing landmarks such as the neck, shoulders, elbows, wrists, and hips. These landmarks serve as explicit indicators of the alignment between human and garment poses, and therefore can be used as guidance to improve the diffusion process. The visualization of human and garment poses is shown in Fig. \ref{fig:pose}.

\subsection{Dynamic Pose Interaction} \label{sec:attention}
To facilitate the dynamic pose interaction, we propose a hierarchical attention module consisting of a pose-aware spatial attention block, a temporal-shift attention block, a cross-attention block, and a pose-aware temporal attention block.

\textbf{Pose-aware Spatial Attention (PASA)}. To capture intricate intra-frame human-garment pose interactions, we inject the features of human and garment poses into the spatial attention layer. In our baseline, the human feature map $\textbf{f}_{h}$ from the main U-Net is concatenated with the corresponding garment feature map $\textbf{f}_{g}$ from the garment U-Net to jointly compute self-attention, which is computed as:
\begin{equation} \label{eq:self-attention}
    \bm{h} = Attn(\psi_q(\textbf{f}), \psi_k(\textbf{f}), \psi_v(\textbf{f})),
\end{equation}
where $\textbf{f}=[\textbf{f}_{h}, \textbf{f}_{g}]$ is the concatenation of human feature
map and garment feature map in the spatial dimension, and $\psi_q, \psi_k, \psi_v$ are linear projections. Afterward, only the half feature from the main U-Net undergoes further computation. This attention layer enables garment features extracted by the garment U-Net to be integrated into the main U-Net. 

To enable the correspondence control with pose guidance, we modify this attention layer by adding the pose embeddings of the human and garment. Formally, given the human pose embedding $\textbf{p}_{h}$ and garment pose embedding $\textbf{p}_{g}$ extracted from the pose encoder, we first concatenate them in spatial dimension $\textbf{p}=[\textbf{p}_{h}, \textbf{p}_{g}]$. Then we utilize a pose adapter to project $\textbf{p}$ into the diffusion feature space. The pose adapter employs two fully connected (FC) layers with an intermediate activation layer. The first FC layer maps the input to a lower-dimensional space, while the second FC layer maps it back to the original dimensional. The pose adapter can be written as:
\begin{equation}
    Adpt(\textbf{p})=\textbf{W}_{up}(GELU(\textbf{W}_{down}(\textbf{p}))),
\end{equation}
where $\textbf{W}_{up}$ and $\textbf{W}_{down}$ are the learnable weight matrix. To preserve the original feature space of diffusion models, we initialize $\textbf{W}_{up}$ with zeros. Subsequently, we add the diffusion feature $\textbf{f}$ and the adapter embedding $Adpt(\textbf{p})$ together and feed them into the self-attention mechanism, leading to the revised form of Eq. \ref{eq:self-attention}:
{\footnotesize
\begin{equation}
    \bm{h} = Attn(\psi_q(\textbf{f}+Adpt(\textbf{p})), \psi_k(\textbf{f}+Adpt(\textbf{p})), \psi_v(\textbf{f}+Adpt(\textbf{p}))).
\end{equation}
}
In this way, when integrating the garment feature into the main U-Net, the feature aggregation process takes into account the alignment of human and garment poses. This implicit consideration results in the garment being deformed to better fit the natural pose of the person.

Existing PoseGuider-style methods \cite{hu2024animate,xu2024tunnel,wang2024gpd} typically achieve pose control by directly adding processed human pose images and noisy latents as inputs to the first layer of the diffusion network, without incorporating pose information into the attention modules. In contrast, our PASA introduces pose information into each layer's attention modules through a pose adapter, enabling more precise and varying degrees of pose control. Moreover, besides human poses, we further inject garment poses into PASA, thus achieving intricate intra-frame human-garment pose alignment.

\textbf{Temporal-shift Attention (TSA)}. In the original SD model, the attention block of the U-Net solely focuses on self-attention within individual frames, overlooking valuable information across frames. Current VVTON techniques predominantly integrate a temporal attention layer at each stage to capture temporal dependencies among video frames. Nevertheless, this approach can introduce a covariate shift in the feature space, potentially undermining the model's generative capacity \cite{zhang2024towards}. While a direct remedy could involve joint space-time 3D attention, this method escalates the complexity of attention calculations quadratically. Drawing inspiration from prior works \cite{lin2019tsm, an2023latent, xing2024simda}, a novel approach involves employing temporal shift operations that utilize a 2D module to amalgamate spatial and temporal information by incorporating neighboring frame data into the current frame. Specifically, in addition to examining tokens within the present frame, a patch-level shifting operation transpires along the temporal dimension, transferring tokens from the preceding $L$ frames to the current frame, thereby constructing a novel latent feature frame. We concatenate the latent feature of the current frame $\textbf{h}$ with the temporally shifted latent feature $\textbf{h}_{shift}$ in the spatial dimension, thus establishing keys and values for subsequent self-attention mechanisms. The temporal-shift attention can be formally written as:
\begin{equation} \label{eq:temporal-shit}
    \bm{\hat{h}} = Attn(\psi_q(\textbf{h}), \psi_k([\textbf{h}, \textbf{h}_{shift}]), \psi_v([\textbf{h}, \textbf{h}_{shift}]).
\end{equation}
It notably reduces the computational load when juxtaposed with 3D attention. Furthermore, it empowers the model to discern short-term relationships between adjacent frames, enhancing temporal coherence during video generation.

\textbf{Cross-Attention (CA)}. In contrast to the original SD model, which employs text embeddings to steer the diffusion process with cross-attention, we have replaced these embeddings with garment image embeddings generated from a CLIP image encoder. These embeddings encapsulate comprehensive global information like color and style, which serve as a valuable supplement to the detailed appearance characteristics provided by the Garment U-Net.

\textbf{Pose-aware Temporal Attention (PATA)}. Although TSA can effectively capture short-term relationships between adjacent frames, it fails to consider the long-term temporal dynamics across the entire video. Previous studies \cite{guo2023animatediff, hu2024animate, fang2024vivid} have introduced a plug-and-play motion module with temporal attention to enhance video smoothness. However, these approaches overlook the dynamic changes in human poses, which are crucial for the VVTON task where garment appearance must adapt to individual poses. To address this limitation, we propose a pose-aware temporal attention. Similar to PASA, our method involves feeding the human pose embedding $\textbf{p}_{h}$ into a pose adapter. This adapted embedding is then element-wise added to the output feature of the cross-attention block, followed by standard temporal attention. Our experiments have demonstrated that this straightforward strategy significantly improves video continuity and enhances the realism of clothing dynamics in motion, as shown in Fig. \ref{fig:experiment_temporal}.

\subsection{Temporal Regularized Attention Loss} \label{sec:loss}
Generally, SD is merely optimized with the mean-squared loss defined in Eq. \ref{eq:ldm}, which treats all regions of the synthesized video equally without emphasizing temporal consistency. Inspired by the idea that self-attention maps within diffusion models capture the structural essence of generated content \cite{cao2023masactrl, jeong2023training, tumanyan2023plug, kwon2024harivo}, we employ the \emph{temporal regularized attention loss}. This loss function is designed to minimize variations in self-attention maps across successive frames:
\begin{equation}
\mathcal{L}_{\text{TRA}} = \sum_{i}^{N} \sum_{j=2}^{T} \gamma_{{i}} | \mathcal{A}^{(j)}_{{i}} - \mathcal{A}^{(j-1)}_{{i}} |,
\vspace{-0.1in}
\end{equation}
where $\mathcal{A}^{(j)}_{{i}}$ represents the self-attention map of the PASA block in the $i$-th layer on $j$-th frame. Specifically, We calculate $L_{TRA}$ only on the last two decoder layers in the main U-Net, with $\gamma_{i}=0.5$ in our context.

\subsection{Training and Inference} \label{sec:train}
\textbf{Training Strategy}. Different from previous multi-stage training approaches \cite{xu2024tunnel, wang2024gpd}, we employ a single-stage training strategy. The loss function is defined as follows:
\begin{equation}
\mathcal{L} = \mathcal{L}_{LDM} + \lambda\mathcal{L}_{TRA},
\end{equation}
where $\lambda$ is the hyper-parameter used to balance the MSE loss and the proposed TRA loss. When training with image datasets, the TSA and PATA blocks are omitted. As for video datasets, we implement an image-video joint training strategy. Initially, we randomly pick $B$ videos from the dataset and extract a single frame randomly from each video for training. At this stage, solely the PASA and CA modules undergo training. Subsequently, in the following phase, another set of $B$ videos is randomly chosen and a sequence of $T$ consecutive frames is extracted from each video for training purposes. Here, the training focuses exclusively on the TSA and PATA modules. This process repeats iteratively until the training concludes. Here, $B$ represents the batch size, and $T$ symbolizes the length of video clips. This training methodology effectively reduces the GPU memory requirements and enhances the speed of convergence.

To tackle discontinuous pose sequences, we also implement a condition dropping strategy during training. Specifically, each pose keypoint has a 0.05 probability of being dropped. When a keypoint of the current frame is dropped, the model is forced to refer to the poses of adjacent frames to infer the correct pose of the current frame. This strategy not only enhances the model's robustness to pose estimation methods but also improves temporal consistency.

\textbf{Inference Strategy}. To improve temporal coherence and smoothness in long videos, we follow \cite{fang2024vivid, wang2024gpd} and implement a sliding window approach during inference. It involves dividing the long video into overlapping segments of a specified length $T$ and conducting inference on each segment. For overlapping frames, the ultimate output is derived by averaging the results obtained from each inference.
\section{Experiments}

\subsection{Datasets}
\textbf{Video Virtual Try-On}. We empirically verify and analyze the effectiveness of our DPIDM on two popular video virtual try-on datasets, VVT \cite{dong2019fw} and ViViD \cite{fang2024vivid}. The VVT dataset contains 791 video clips with a resolution of 256 $\times$ 192. Following the official settings\cite{dong2019fw, xu2024tunnel}, we partition it into a training set of 661 clips and a test set of 130 clips. However, the VVT dataset exhibits limited diversity in terms of garment varieties and human actions. ViViD, a recent and challenging dataset, includes 9,700 video-image pairs at a resolution of 832 $\times$ 624. It classifies garments into three categories: upper-body, lower-body, and dresses. The dataset is divided into 7,759 videos for training and 1,941 videos for testing. It is worth noting that due to inaccuracies in the provided cloth-agnostic masks for individuals in a back-facing posture, we excluded video segments depicting such scenarios from the original footage. Consequently, only the remaining segments were utilized for testing.

\textbf{Image Virtual Try-On}. To further validate the effectiveness of our method, we carried out additional experiments on the widely recognized image-based virtual try-on dataset, VITON-HD \cite{choi2021viton}. This dataset comprises 13,679 pairs of upper-body model and garment images, out of which 2,032 pairs were specifically reserved for testing.

\textbf{Evaluation Metrics}. For image datasets, we use Structural Similarity (SSIM) \cite{wang2004image} and Learned Perceptual Image Patch Similarity (LPIPS) \cite{zhang2018unreasonable} to measure the similarity between the generated image and the ground-truth. Additionally, Fréchet Inception Distance(FID) \cite{heusel2017gans} and Kernel Inception Distance(KID) \cite{binkowski2018demystifying} are employed to measure the quality and realism of the generated images. For video datasets, we further utilize Video Fréchet Inception Distance (VFID) \cite{unterthiner2018towards} to evaluate spatiotemporal consistency. Two CNN backbones, I3D \cite{carreira2017quo} and 3D-ResNeXt101 \cite{hara2018can}, are adopted as feature extractors for VFID.

\subsection{Implementation Details}
We initialize the main U-Net with the pre-trained SD v1.5 inpainting model and the garment U-Net with SD v1.5. The pose estimator is trained offline using YOLOv8 \cite{redmon2016you}. The pose encoder incorporates four convolutional layers to align the pose image with the noise latent at the same resolution. During training, all data is resized to a uniform resolution of 512 × 384. To improve model resilience, we employ data augmentation by randomly flipping images horizontally with a $50\%$ probability. We utilize the Adam optimizer \cite{kingma2014adam} with a consistent learning rate of $2\times10^{-5}$. All the experiments are conducted using 16 NVIDIA A100 GPUs. Sequences of 24 frames are sampled, and the batch size $B$ is set to 32. The model is trained for 80,000 iterations. The hyper-parameter $\lambda$ is set to $10^{-3}$ for video datasets and 0 for image datasets. During inference, we utilize the DDIM \cite{song2020denoising} sampler and set the classifier-free guidance scale to 1.5.



\subsection{Quantitative Results}
\begin{table}[t]
\small
\renewcommand{\arraystretch}{1}
\caption{Comparison results on the VVT dataset. VFID$_{I}$ and VFID$_{R}$ represent VFID$_{I3D}$ and VFID$_{ResNeXt}$ respectively. $\uparrow$ denotes higher is better, while $\downarrow$ indicates lower is better.}
\setlength\tabcolsep{4pt}
\centering
\begin{tabular}{l|cccccc}
\toprule 
\textbf{Method} & \textbf{SSIM$\uparrow$} & \textbf{LPIPS$\downarrow$} & \textbf{VFID$_{I}$$\downarrow$}  & \textbf{VFID$_{R}$$\downarrow$}\\
\midrule
CP-VTON \cite{wang2018toward}           & 0.459 & 0.535 & 6.361 & 12.10\\
FW-GAN \cite{dong2019fw}                & 0.675 & 0.283 & 8.019 & 12.15\\
PBAFN \cite{ge2021parser}               & 0.870 & 0.157 & 4.516 & 8.690\\
ClothFormer \cite{jiang2022clothformer} & 0.921 & 0.081 & 3.967 & 5.048\\
\midrule
LaDI-VTON \cite{morelli2023ladi}        & 0.878 & 0.190 & 5.880 & - \\
StableVITON \cite{kim2023stableviton}   & 0.876 & 0.076 & 4.021 & 5.076\\
TunnelTry-on \cite{xu2024tunnel}       & 0.913 & \underline{0.054} & 3.345 & 4.614\\
ViViD \cite{fang2024vivid}              & \textbf{0.949} & 0.068 & 3.405 & 5.074\\
VITON-DiT \cite{zheng2024viton-dit}     & 0.896 & 0.080 & 2.498 & \underline{0.187}\\
GPD-VVTO \cite{wang2024gpd}             & 0.928 & 0.056 & \underline{1.280} & - \\
\textbf{DPIDM (Ours)} & \underline{0.930} & \textbf{0.041} & \textbf{0.506} & \textbf{0.047}\\

\bottomrule 
\end{tabular}
\label{tab:vvt}
\vspace{-0.1in}
\end{table}

\begin{table}[t]
\small
\renewcommand{\arraystretch}{1}
\caption{Comparison results on the ViViD dataset.}
\setlength\tabcolsep{4pt}
\centering
\begin{tabular}{l|ccccc}
\toprule 
\textbf{Method} & \textbf{SSIM$\uparrow$} & \textbf{LPIPS$\downarrow$} & \textbf{VFID$_{I}$$\downarrow$}  & \textbf{VFID$_{R}$$\downarrow$}\\
\midrule
LaDI-VTON \cite{morelli2023ladi}  &0.824  &0.164  &8.283  &1.185 \\
CAT-DM \cite{zeng2024cat}  &0.826  &0.162  &9.354  &2.139 \\
ViViD \cite{fang2024vivid}        &\underline{0.846} &\underline{0.118} &\underline{1.894} &\underline{0.870} \\
\textbf{DPIDM (Ours)} &\textbf{0.883} &\textbf{0.081} &\textbf{0.488} &\textbf{0.090} \\
\bottomrule 
\end{tabular}
\label{tab:vivid}
\vspace{-0.1in}
\end{table}

\begin{table}[t]
\small
\renewcommand{\arraystretch}{1}
\caption{Quantitative results of image-based virtual try-on task on the VITON-HD dataset. FID$_{u}$/KID$_{u}$ stands for the FID/KID score in unpaired setting. Note that the KID score is multiplied by 1000.}
\setlength\tabcolsep{4pt}
\centering
\begin{tabular}{l|ccccc}
\toprule 
\textbf{Method} & \textbf{SSIM$\uparrow$} & \textbf{LPIPS$\downarrow$} & \textbf{FID$_{u}$$\downarrow$}  & \textbf{KID$_{u}$$\downarrow$}\\
\midrule
VITON-HD \cite{choi2021viton} & 0.862 & 0.117 & 12.12 & 3.23 \\
HR-VITON \cite{lee2022high} & 0.876 & 0.096 & 12.31 & 3.81 \\
GP-VTON \cite{xie2023gp} & 0.890 & 0.085 & 9.82 & 1.42 \\
\midrule
LaDI-VTON \cite{morelli2023ladi} & 0.875 & 0.091 & 9.32 & 1.55 \\
WearAnyWay \cite{chen2024wear} & 0.877 & 0.078 & \underline{8.16} & 0.78 \\
DCI-VTON \cite{gou2023taming} & 0.890 & 0.072 & 8.77 & 0.89 \\
StableVITON \cite{kim2023stableviton} & 0.878 & 0.075 & 9.43 & 1.54 \\
CAT-DM \cite{zeng2024cat}  &0.877  &0.080  &8.93  &1.37 \\
IDM \cite{choi2024improving} & 0.881 & 0.078 & 8.60 & \underline{0.55} \\
GPD-VVTO \cite{wang2024gpd} & \underline{0.891} & \underline{0.070} & 8.57 & 0.78 \\
\textbf{DPIDM (Ours)} & \textbf{0.893} & \textbf{0.067} & \textbf{8.15} & \textbf{0.32} &\\
\bottomrule 
\end{tabular}
\label{tab:viton}
\vspace{-0.15in}
\end{table}

\textbf{Video Virtual Try-On}. We compare our DPIDM with a series of state-of-the-art virtual try-on methods, which can be grouped into two directions: GAN-based methods \cite{wang2018toward,dong2019fw,ge2021parser,jiang2022clothformer} and Diffusion-based methods \cite{morelli2023ladi,kim2023stableviton,xu2024tunnel,fang2024vivid,zheng2024viton-dit,zeng2024cat,wang2024gpd}. Table \ref{tab:vvt} summarizes the performance comparisons on the VVT dataset. Notably, our DPIDM consistently outperforms other state-of-the-art methods across LPIPS and VFID metrics. Although ViViD \cite{fang2024vivid} attains a higher SSIM score, its advantage is expected due to the utilization of a significantly larger volume of high-resolution videos during training, approximately six times more than the VVT dataset. Despite this, our DPIDM notably surpasses ViViD in the VFID metric, showcasing substantial enhancements in video coherence while preserving garment visual authenticity. 
ClothFormer improves FW-GAN by introducing an appearance-flow tracking module to ensure temporal consistency in garment warping. TunnelTry-on and ViViD further boost the performances by leveraging the capabilities of pre-trained diffusion models. Additionally, GPD-VVTO introduces garment-aware temporal attention to enhance temporal consistency, leading to a clear VFID score boost. However, existing approaches often neglect crucial spatiotemporal pose interactions between individuals and garments. In contrast, our DPIDM effectively models both intra-frame human-garment pose interactions and long-term human pose dynamics across frames through pose-aware attention mechanisms, leading to superior results in the VVTON task. Specifically, our DPIDM achieves a VFID$_{I}$ score of 0.506, marking a substantial 60.5\% relative improvement over the top competitor GPD-VVTO. Note that GPD-VVTO employs SD v2.1 for model initialization, whereas we choose SD v1.5 for fewer parameters.

Table \ref{tab:vivid} presents performance comparisons on the ViViD dataset. As observed with VVT, DPIDM consistently outperforms other VVTON methods across all metrics. which again evinces the pivotal merit of the pose-aware guidance for preserving spatiotemporal consistency in the generated videos. Particularly, DPIDM demonstrates a substantial relative improvement of 74.2\% over ViViD in VFID$_{I}$. 

\textbf{Image Virtual Try-On}. Table \ref{tab:viton} displays the quantitative results on the VITON-HD \cite{choi2021viton}. Following the official settings, we utilize SSIM and LPIPS metrics for the paired setting, and FID and KID for the unpaired setting. Our DPIDM consistently outperforms other methods across all metrics, showcasing its ability to produce authentic and lifelike images while maintaining the integrity of the initial garment structure. It is noteworthy that diffusion-based approaches exhibit lower FID and KID scores compared to GAN-based methods, suggesting that diffusion models excel in generating images with heightened fidelity.

\begin{figure}[t]
\centering
    \setlength{\abovecaptionskip}{0pt}
    \includegraphics[width=0.475\textwidth]{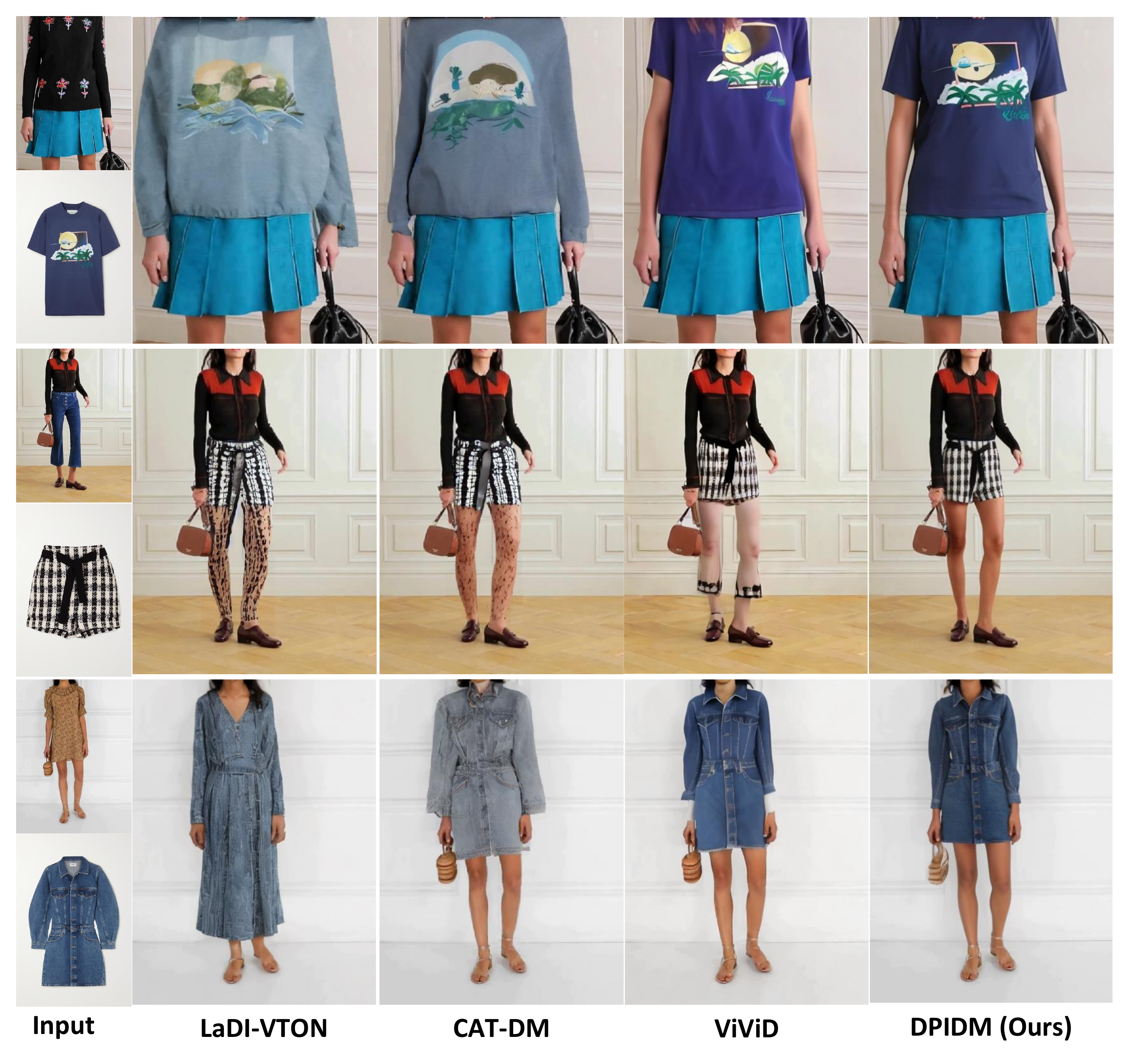}
    \caption{Qualitative comparison on the ViViD dataset. Our DPIDM excels in seamlessly integrating the garment with the wearer and maintaining the visual integrity of the garment.
    }
    \label{fig:spatial}
\vspace{-0.1in}
\end{figure}

\begin{figure*}[ht]
\centering
    \setlength{\abovecaptionskip}{0pt}
    \includegraphics[width=0.95\linewidth]{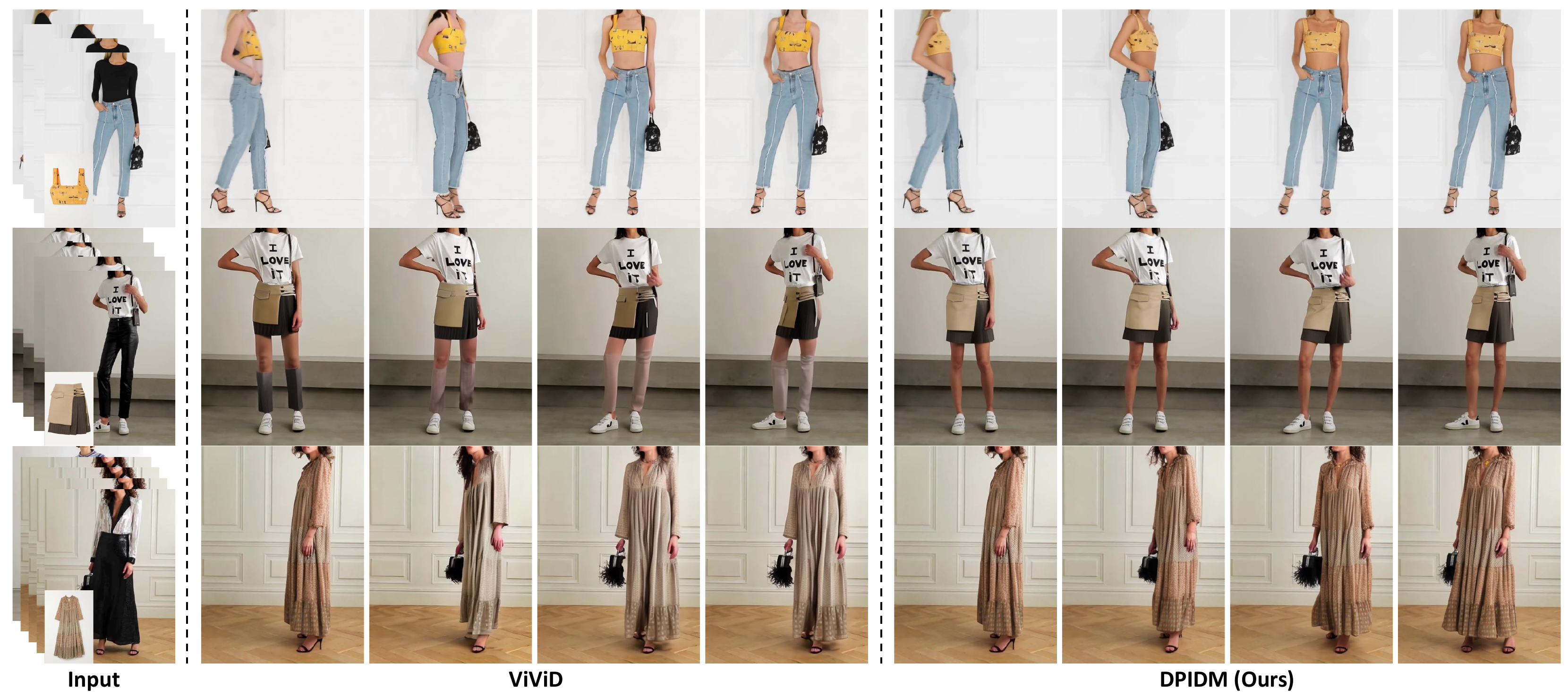}
    \caption{Qualitative comparison on the ViViD dataset. Our DPIDM maintains temporal consistency even during substantial movements.
    }
    \label{ablation1}
\label{fig:experiment_temporal}
\vspace{-0.15in}
\end{figure*}

\subsection{Qualitative Results}
We conduct qualitative analysis from two aspects: visual authenticity of the garment and temporal consistency. When evaluating visual authenticity, we compare results from several virtual try-on methods using individual video frames, as illustrated in Figure \ref{fig:spatial}. In the first row, LaDI-VTON and CAT-DM change the style of the target garment and introduce color discrepancies. While ViViD achieves a natural fit, it struggles to maintain the original patterns, leading to their displacement on the garment. In contrast, our method excels in seamlessly blending the garment with the wearer while preserving intricate fabric patterns. Moving to the second row with a pair of striped shorts, competitors maintain the garment's style but exhibit noticeable artifacts around the legs. Conversely, our method adeptly addresses these issues by naturally supplementing missing skin tones while upholding the garment's visual integrity. This trend continues in the subsequent example with a dress in the third row, highlighting the effectiveness of DPIDM. To further evaluate temporal consistency, we extract four frames at equal intervals from the generated video and present them in Fig. \ref{fig:experiment_temporal}. While ViViD achieves a natural fit in individual frames, it struggles to maintain garment details and temporal consistency across frames, especially during person movements. In contrast, our DPIDM consistently delivers high-quality results that preserve garment details and ensure temporal consistency without introducing artifacts, even during actions involving significant movements.

\subsection{Ablation Studies}

\begin{table}[!t]
\renewcommand{\arraystretch}{1.1}
\caption{Performance contribution of each component in DPIDM.}
\footnotesize
\setlength\tabcolsep{4pt}
\centering
\begin{tabular}{c|ccc|cccc}
\toprule 
{\emph{id}} & {PAA} & {TSA} & {TRA} & \textbf{SSIM$\uparrow$} & \textbf{LPIPS$\downarrow$} & \textbf{VFID$_{I}$$\downarrow$}  & \textbf{VFID$_{R}$$\downarrow$} \\
\midrule
(a) & \textcolor{lightgray}{\ding{55}} & \textcolor{lightgray}{\ding{55}} & \textcolor{lightgray}{\ding{55}} & 0.893 & 0.084 & 3.451 &2.435 \\
(b) & \ding{51}                        & \textcolor{lightgray}{\ding{55}} & \textcolor{lightgray}{\ding{55}} & 0.925 & 0.050 & 1.068 & 0.153 \\
(c) & \ding{51}                        & \ding{51}                        & \textcolor{lightgray}{\ding{55}} & 0.929 & 0.043 & 0.721 & 0.075\\
(d) & \ding{51}                        & \ding{51}                        & \ding{51}                        & \textbf{0.930} & \textbf{0.041} & \textbf{0.506} & \textbf{0.047}\\
\bottomrule 
\end{tabular}
\label{tab:ablation}
\vspace{-0.1in}
\end{table}

In this study, we analyze the impact of each design element within DPIDM on the overall performance using the VVT dataset. As shown in Table \ref{tab:ablation}, the proposed elements are denoted as follows: \emph{PAA} refers to pose-aware attention, \emph{TSA} to temporal-shift attention, and \emph{TRA} to temporal regularized attention loss. We start with a basic model (a) that directly inserts standard temporal attention after each stage of the main U-Net. Subsequently, in (b), we substitute the standard spatial/temporal attention with our proposed pose-aware spatial/temporal attention. Notably, the introduction of \emph{PAA} yields a significant performance enhancement over (a), surpassing even the state-of-the-art methods detailed in Table \ref{tab:vvt}. This not only validates the efficacy of our proposed pose-aware spatial/temporal attention but also underscores the critical role of integrating spatiotemporal pose interactions into the VVTON task. Furthermore, we introduce the temporal-shift attention and the temporal regularized attention loss in (c) and (d) respectively. While \emph{TSA} and \emph{TRA} did not result in significant enhancements in the SSIM and LPIPS metrics, they notably improved the VFID metric, highlighting their effectiveness in enhancing video consistency. Collectively, the innovative modules introduced in DPIDM enhance the visual integrity of the garment and improve the temporal consistency of the generated videos.

\section{Conclusion}
In this work, we have presented Dynamic Pose Interaction Diffusion Models (DPIDM), a novel approach that capitalizes on spatial and temporal pose interactions to enhance video virtual try-on. Particularly, we study the problem from the viewpoint of employing intra-frame human-garment pose interactions and long-term human pose dynamics across frames. To verify our claim, we develop a hierarchical pose-aware attention module to integrate human and garment pose features into spatial and temporal attention, enhancing the temporal coherence of the generated videos while preserving the visual fidelity of the garments. Additionally, a new temporal regularized attention loss is devised to bolster consistency across successive frames. Through extensive experiments, we demonstrate that our DPIDM surpasses existing state-of-the-art methods on image-based and video-based virtual try-on datasets.

\section*{Acknowledgments} This work was supported in part by the Beijing Municipal Science and Technology Project No. Z241100001324002 and Beijing Nova Program No. 20240484681.
{
    \small
    \bibliographystyle{ieeenat_fullname}
    \bibliography{main}
}


\end{document}